\documentclass[a4paper,10pt]{article}
\usepackage{graphicx,wrapfig}
\usepackage{amssymb,amsmath,amsthm}
\usepackage{xcolor}
\usepackage{algorithm}
\usepackage{algorithmic}
\newtheorem{definition}{Definition}
\newtheorem{proposition}{Proposition}
\newtheorem{theorem}{Theorem}
\DeclareMathOperator{\argmin}{argmin}
\newcommand{\uargmin}[1]{\underset{#1}{\argmin}\;}
\DeclareFontFamily{U}{mathx}{\hyphenchar\font45}
\DeclareFontShape{U}{mathx}{m}{n}{
      <5> <6> <7> <8> <9> <10>
      <10.95> <12> <14.4> <17.28> <20.74> <24.88>
      mathx10
      }{}
\DeclareSymbolFont{mathx}{U}{mathx}{m}{n}
\DeclareFontSubstitution{U}{mathx}{m}{n}
\DeclareMathAccent{\widebar}{0}{mathx}{"73}      
\newcommand{\R}{\mathbb{R}}
\newcommand{\E}{\mathbb{E}}
\newcommand{\A}{\mathcal{A}}

\newcommand{\M}{\mathcal{M}}
\renewcommand{\P}{\mathbb{P}}
\newcommand{\X}{\mathcal{X}}

\newcommand{\ones}{\mathbf{1}}

\newcommand{\bfx}{\mathbf{x}}
\newcommand{\bftheta}{\boldsymbol\theta}
\newcommand{\bfomega}{\boldsymbol\omega}
\newcommand{\bfupsilon}{\boldsymbol\upsilon}
\newcommand{\bfgamma}{\boldsymbol\gamma}
\newcommand{\bfeta}{\boldsymbol\eta}
\newcommand{\bfpi}{\boldsymbol\pi}
\newcommand{\bfmu}{\boldsymbol\mu}

\begin{document}

\title{A unified framework for hard and soft clustering with regularized optimal transport}

\date{}
\author{Jean-Frédéric Diebold\footnote{Safran Tech} \and
Nicolas Papadakis\footnote{Univ. Bordeaux, CNRS, INRIA, Bordeaux INP, IMB, UMR 5251, F-33400 Talence, France}  \and
Arnaud Dessein\footnote{BackMarket} \and Charles-Alban Deledalle\footnote{Brain Corp}}
%


%
\maketitle              % typeset the header of the contribution
\begin{abstract}
In this paper, we formulate the problem of inferring a Finite Mixture Model from discrete data as an optimal transport problem with entropic regularization of parameter $\lambda\geq 0$. Our method unifies hard and soft clustering, the  Expectation-Maximization (EM) algorithm being exactly recovered for $\lambda=1$. The family of clustering algorithm we propose rely on the resolution of nonconvex problems using alternating minimization. We study the convergence property of our generalized $\lambda-$EM algorithms and show that each step in the minimization process has a closed form solution when inferring finite mixture models of exponential families. Experiments highlight the benefits of taking a parameter $\lambda>1$ to improve the inference performance and $\lambda\to 0$ for classification.
\end{abstract}

\section{Introduction}
The inference of a probabilistic parametric model from a set of realizations consists in estimating parameters allowing an explanation or even a generalization of these realizations. 
The inference of finite mixtures of Gaussian distributions, or more generally of exponential families, is thus widely used in data science tasks such as video modeling~\cite{greenspan2004probabilistic}, image segmentation~\cite{Carson2002}, image restoration~\cite{zoran2011learning}, audio source separation~\cite{Benaroya2006}, speech/music discrimination~\cite{Scheirer1997} and musical style classification~\cite{Tzanetakis2002}. 
To estimate model parameters, a classical unsupervised approach is maximizing the likelihood. 
The likelihood is computed as the probability (or probability density) of realizations $\{x_1,\ldots,x_n\}$ along a model depending on parameters $\eta$. In many cases, such as for mixtures of exponential families, maximizing the likelihood with respect to the model parameters is not straightforward. Then, enriching the observations $x$ with hidden latent states $\pi$ allows to maximize the likelihood of the equivalent model in $(x,\pi)$ in a more convenient way. As an example, when considering mixtures of exponential families, if $\pi$ describes the belonging of realizations $x$ to their respective cluster, the inference of the law of each {\em cluster} (i.e. group of the mixture) can be done independently and admits a closed form expression. %The pair $(x,\pi)$ is called complete information.

Several {\em hard} and {\em soft} clustering algorithms rely on the complete information formulation in $(x,\pi)$ to iteratively infer parameters that simultaneously increase the likelihood of the observations $x$ with respect to the model. One example is the k-means (KM) algorithm, which assigns each realization $x_i$ to a single centroid (known as hard clustering) before updating its position. The Expectation-Maximization (EM) algorithm, proposed  in~\cite{dempster1977maximum}, is a soft clustering algorithm that distributes realizations $x_i$ to all clusters $j$ according to an a posteriori law $\pi_{ij}$ which is updated at each iteration to increase the global likelihood with respect to $x$ of the parameterized model. 
Such a soft strategy is meant to be more robust than a hard one so as to prevent from errors to occur and propagate when there is possible ambiguity in the assignment, though both approaches can be competitive in practice depending on the task at hand and the dataset.

\paragraph{Contributions and outline.}
The main contribution of this paper is to bridge the gap between hard and soft clustering by introducing a general formulation of statistical inference based on Regularized Optimal Transport (ROT).
We  cast the inference of finite mixture models as a ROT  problem with an entropic regularization of  parameter $\lambda\geq 0$. 
Solving the ROT problem amounts to minimizing a nonconvex functional. We propose to optimize this functional with an alternating minimization method. We show that each minimization step admits a closed form solution for exponential family models, and prove that accumulation points of the alternating algorithm are stationary points of the functional.
Finally, we demonstrate that k-means and EM are recovered as specific cases of our algorithm for $\lambda=0$ and $\lambda=1$. Contrary to existing methods for statistical inference through optimal transport~\cite{mena2020sinkhorn}, the computational burden of our approach does not increase with data dimension. 
We finally  illustrate the interest of taking a parameter $\lambda>1$ for the the inference of $2D$ Gaussian Mixture Models and recovering  hard clustering with $\lambda=0$ for classification.

After a review of related works in section~\ref{sec:related}, we introduce the regularized optimal transport problem for generalizing the EM algorithm in section~\ref{sec:formulation}. 
In section~\ref{sec:optim}, we analyze mathematically the convergence properties of the numerical algorithm. In section~\ref{sec:exp_fam}, we specify the algorithm  for mixtures of exponential families, and analyse the connection between our generalized algorithm and the maximization of the likelihood.  Finally, we provide  in Section~\ref{sec:expe} experiments illustrating the robustness properties brought by the regularization parameter on the inferred solutions.

\section{Related Work}\label{sec:related}

\paragraph{Hard clustering and $k$-Means}
The classical $k$-Means (KM) algorithm~\cite{Lloyd1982} is an iterative relocation scheme that seeks to minimize the average Euclidean distance of points to their respective cluster center. One major improvement of KM is the KM++ algorithm~\cite{Arthur2007}, that  proposes a tailored  initialization  of clusters  with guaranteed performance bounds. 

\paragraph{Soft clustering and Expectation-Maximization (EM) in Exponential Family Mixture Models}
Broadly speaking, EM is a parameter estimation method that infers the parameters and weights of the $k$ Gaussian components of the mixture, by trying to reach the maximum likelihood (ML) of the observations. In general, EM is initialized using the estimated solutions from KM++, where the $k$ initial parameters are obtained by Maximum Likelihood  in each separate cluster and the weights are proportional to the number of assigned points. The likelihood of the data is non-decreasing at each iteration. Further, if there exists at least one local maximum for the likelihood function, then the algorithm will converge to a local maximum of the likelihood. The framework has then been generalized to exponential families in~\cite{Banerjee2005}.

\paragraph{Relations between Hard and Soft Clustering or Estimation}
KM can be interpreted as a limit case of EM for isotropic Gaussian mixtures with variance $\sigma \to 0$.  \cite{Kearns1997} give further light on the relationship using an information-theoretic analysis of hard KM and soft EM assignments in clustering. 
We note that the Bregman hard clustering algorithm~\cite{Banerjee2005} is a limiting case of the above soft clustering algorithm. 
%For every convex function f and positive constant b, bf is also a convex function with the corresponding Bregman divergence. 
In the limit, the posterior probabilities in the E-step take values in $\{0, 1\}$ and the E and M steps of the soft clustering algorithm reduce to the assignment and re-estimation steps of the hard clustering algorithm.
KM can be used as initialization for a EM, or in more evolved hard clustering algorithms such as $k$\nobreakdash-MLE~\cite{Nielsen2012}.

\paragraph{Statistical Inference by Minimizing Information Divergences}
There exists an important literature on statistical inference by minimizing discrepancy measures between observed data and model, so as to obtain more robust estimators and tests while maintaining sufficient efficiency~\cite{Eguchi1983,Basu1998,Eguchi2001,Mihoko2002,Pardo2006,Eguchi2009,Basu2011}. These works mainly focus on information divergences, with known relation between MLE and KL or Bregman divergence for (curved) exponential families.

\paragraph{Comparison of Exponential Family Mixture Models by Optimal Transport}
Optimal Transport (OT) defines a family of distance between distributions.
It  consists in estimating a map transferring a source distribution onto a target one that minimizes a given cost of displacement. The OT distance (also known as the Wasserstein distance
or the Earth Mover distance) has been shown to produce state
of the art results for the comparison of Gaussian Mixture Models (GMMs)~\cite{Rubner2000,chen2018optimal} and generalized to exponential families in~\cite{Zhang2009}. The OT problem, that can be solved with linear programming,  has a prohibitive computational cost for large scale problems. 
For this reason, Regularized  Optimal Transport (ROT)  models have received a lot of attention in Machine Learning and Imaging~\cite{peyre2019computational}.

\paragraph{Clustering and Statistical Inference by  OT minimization}
Using the concept of Wasserstein barycenters~\cite{Carlier_wasserstein_barycenter}, there exists many clustering algorithms based on OT~\cite{irpino2006new,del2016robust,marti2016optimal,2016arXiv160500513I,ho2017multilevel,ye2015fast,zhuang2022wasserstein}. A  generalization of KM++ with Optimal Transport cost has also been proposed in~\cite{WKMEAN}.

Since the theoretical work in~\cite{BASSETTI20061298}, the  statistical inference by minimizing OT  distances, also known as the Minimum Kantorovitch Estimator problem~\cite{bernton2019parameter,peyre2019computational}, has  been increasingly studied~\cite{frogner2020approximate,blanchet2021sample,lambert2022variational,yi2023sliced}.~Numerical algorithms have also been proposed  through Boltzmann Machines~\cite{NIPS2016_6248}. 
As the parameter estimation problem is non convex and potentially non differentiable, these methods fail at obtaining good parameterizations  for large scale problems. 
Instead of estimating parameters of prior laws, Wasserstein  Generative Adversarial Networks~\cite{arjovsky2017wasserstein} design   generators that provide samples  respecting the data distribution according to regularized or approximate OT distances~~\cite{bousquet2017optimal,genevay2018learning,adler2018banach,houdard2022gradient}.

There exist no general framework for the use of OT in ``traditional'' algorithms such as KM, or EM. In this paper, we fill this gap and study numerical algorithms considering statistical inference based on ROT.
We nevertheless mention the method of~\cite{mena2020sinkhorn} that learns GMMs by solving an OT problem. Notice that this competing approach is limited to cases where the weights of the mixture components are known and fixed. It also involves the resolution of a ROT problem on a discretized grid at each iteration, thus limiting the experiments to 1D examples or rough discretization. On the other hand, our algorithm admits closed form expressions for each step, and the computational complexity is only  related to the number of clusters and independent to the data dimension.

\section{Parameter Estimation in Finite Mixture Models with Regularized Optimal Transport}\label{sec:formulation}

In this section, we define our inference framework.  We introduce the notations relative to clustering in section~\ref{ssec:notations} and finite mixture models in section~\ref{ssec:fmm}. The formulation of statistical inference as an optimal transport problem is proposed in section~\ref{ssec:pb}. This corresponds to estimating a low dimensional regularized optimal transport (ROT) map between the sample points distribution and the weighted distribution of cluster parameters.  In section~\ref{ssec:relaxed}, we show that the constraint on the marginal relative to the cluster weights can be removed, and we propose to recast the problem as a simplified nonconvex functional to optimize.

\subsection{Hard and soft Clustering}~\label{ssec:notations}

Broadly speaking, hard and soft clustering methods suppose that a series of $n$ observations $(x_i)_{i = 1}^n$ from a space $\X$ are issued from a mixture of $k$ components $(\X_j)_{j = 1}^k$ called clusters. Typically, the belonging of each point $x_i$ to each cluster $\X_j$ is represented by a membership coupling $\pi_{ij}$. In general, the weights $\omega_j$ of the clusters $\X_j$ are estimated through the proportion of points that are assigned to them. The clusters $\X_j$ are also represented by individual models $p_{\eta_j}$. A generic scheme to optimize all variables  is thus to update the memberships $\pi_{ij}$, weights $\omega_j$ and model parameters $\eta_j$ in turn until convergence~\cite{Nock2006}. The difference between hard and soft clustering then lies in the computation of memberships $\pi_{ij}$ for the assignment of points to the clusters. In hard clustering such as KM, the memberships $\pi_{ij} \in \{0, 1\}$ are binary so that each point $x_i$ belongs to exactly one cluster $\X_j$. The clustering thus produces a true partition of the data since the clusters are mutually disjoint. In soft clustering such as EM, the memberships $\pi_{ij} \in [0, 1]$ are relaxed so that each point $x_i$ 
is given a degree of likeliness to belong to the respective clusters $\X_j$.  
Hence, the obtained clusters can all overlap, although it is always possible to obtain a true partition by selecting the cluster with highest membership per point. 

\subsection{Finite Mixture Models}\label{ssec:fmm}

Let $\P = \{P_{\eta}\}_{\eta \in \eta}$ be a parametric model on a measurable space $(\X, \A)$. The finite mixture model with $k$ components from $\P$ is the model $\M = \{P_{\bfomega, \bfeta}\}_{\bfomega \in \Sigma_k, \bfeta \in \Gamma^k}$ on $(\X, \A)$ whose probability measures can be expressed as 
%\begin{equation}
$ P_{\bfomega, \bfeta} = \sum_{j = 1}^k \omega_j P_{\eta_j}$, 
%\end{equation}
where $\bfeta = (\eta_j)_{j = 1}^k \in \Gamma^k$ and $\bfomega = (\omega_j)_{j = 1}^k \in \Sigma_k:=\{\bfomega \in \R_+^k \colon \sum _{j = 1}^k \omega_j = 1\}$  are the  parameters and weights of the $k$ components.

We focus on a dominated and homogeneous model $\P$. In other words, all distributions from $\P$ admit probability densities $p_{\eta}$ with respect to some common $\sigma$\nobreakdash-finite measure $\mu$, and these densities share the same support $\X$. Hence we consider the likelihood of the observations on  $\X$. This framework also extends to probability densities $p_{\bfomega, \bfeta}$ for all distributions from $\M$, expressed on $\X$ as %follows:
%\begin{equation}
$ p_{\bfomega, \bfeta}(x) = \sum_{j = 1}^k \omega_j p_{\eta_j}(x)$. % \enspace.
%\end{equation}

\subsection{General Problem}\label{ssec:pb}
Let $\bfx = (x_i)_{i = 1}^n \in \X^n$ be composed of i.i.d. random variables that are distributed according to an unknown probability measure $p_{\widebar{\bfomega}, \widebar{\bfeta}}$ from $\M$. We call $p_{\widebar{\bfomega}, \widebar{\bfeta}}$ the true distribution. We consider the inference of the weights $\bfomega \in \Sigma_k$ and  ${\bfeta} \in \Gamma^k$ on the basis of the sample observations $\bfx$. 
The dataset $\bfx$ can be expressed as an empirical distribution $p_{\bfupsilon}$ on $\X$ made of the sum of Dirac masses located at points $x_i$ with some arbitrary weights $\bfupsilon = (\upsilon_i)_{i = 1}^n \in \Sigma_n$ chosen by the user:
\begin{equation}
p_{\bfupsilon} = \sum_{i = 1}^n \upsilon_i \delta_{x_i} \enspace.
\end{equation}
The weights are typically taken uniform equal to $1 / n$, though non-uniform weights $\upsilon_i$ can also be used to put more or less emphasis on the respective observations $x_i$ in the estimation process.

We then introduce a second measure for defining the optimal transport problem. This measure is the weighted sum of Dirac localized along parameters $\bfeta$:
\begin{equation}
q_{\bfomega,\bfeta}  = \sum_{j = 1}^k \omega_j \delta_{\eta_j}\enspace.
\end{equation}
From the perspective of optimal transport, we consider the transportation between the distribution $p_{\bfupsilon}$ that is observed and the distribution of clusters $q_{\bfomega,\bfeta} $ to fit. This gives a coupling between observations and source clusters. Introducing a cost matrix $\bfgamma \in \R^{n \times k}$ between pairwise components and the convex and differentiable entropy function $H:\bfpi\in \R^{n \times k}\to \sum_{ij}\pi_{ij}(\log(\pi_{ij})-1)\in\R$ with penalty $\lambda \geq 0$ on transport plans, the ROT between the two distributions reads:
\begin{equation}
\label{eq:rot}
d(p_{\bfupsilon}, q_{\bfomega,\bfeta}) = \inf_{\bfpi \in \Pi(\bfupsilon, \bfomega)} \langle \bfgamma, \bfpi \rangle + \lambda H(\bfpi) \enspace,
\end{equation}
where the transport polytope is defined by:
\begin{equation}
\Pi(\bfupsilon, \bfomega) = \{\bfpi \in \R_+^{n \times k} \colon \bfpi \ones_k = \bfupsilon, \bfpi^\top \ones_n = \bfomega\} \enspace.
\end{equation}
This entropic ROT between discrete distributions is known as the Sinkhorn distance~\cite{cuturi2013sinkhorn}. As discussed in Appendix~\ref{app:gen}, the framework can be extended to other strongly convex regularizations.

We aim at identifying the coupling that maximizes the complete information  log-likelihood in $(\bfx,\bfpi)$. To that end, we introduce the following cost matrix:
\begin{equation}\label{eq:cost}
\gamma_{ij}(\bfomega, \bfeta) = -%\upsilon_i 
\log(\omega_j p_{\eta_j}(x_i)) \enspace.
\end{equation}
Intuitively, the more plausible the observation $i$ is in relation to the component $j$, the less expensive it is to transport the underlying mass. Notice that the cost matrix $\bfgamma$  depends on the variables of the problem. The cost matrix will thus be updated during the optimization process, as it is done in the context of ground cost learning for optimal transport based generative modeling~\cite{genevay2018learning}.

The estimation of the mixture model amounts to finding the weights and parameters that minimize the total transportation cost $d_{\bfomega, \bfeta}(p_{\bfupsilon}, q_{\bfomega, \bfeta})$, that is:
\begin{equation}
\label{generalproblem}
\inf_{\substack{\bfomega \in \Sigma_k\\ \bfeta \in \Gamma^k}} \, \inf_{\bfpi \in \Pi(\bfupsilon, \bfomega)} \lambda H(\bfpi)-\sum_{i = 1}^n \sum_{j = 1}^k \pi_{ij} %\upsilon_i 
\log(\omega_j p_{\eta_j}(x_i))  \;.
\end{equation}
Without any further assumptions, there is no theoretical guarantee that the outer infimum in~\eqref{generalproblem} is actually attained, nor that it is unique. 
The objective~\eqref{generalproblem} being nonconvex, reaching a global optima is not possible. We rather target a critical point with an algorithm that iteratively decreases the objective function

\subsection{Proposed model with Transport Plan Relaxation}\label{ssec:relaxed}

An intuitive strategy to solve the general problem~\eqref{generalproblem} is to perform an alternate optimization on the different variables $\bfomega, \bfeta, \bfpi$. However, for fixed weights $\bfomega$ and parameters $\bfeta$, updating the transport plan $\bfpi$ requires solving a full regularized optimal transport problem. This gets costly as the optimization is repeated through the alternate updates. We thus formulate the relaxed problem \eqref{problem}, where we replace the constraint $\bfpi \in \Pi(\bfupsilon, \bfomega)$ with $\bfpi \in \Pi(\bfupsilon):= \{\bfpi \in \R_+^{n \times k} \colon \bfpi \ones_k = \bfupsilon\}$:
\begin{equation}
\label{problem}
\inf_{\substack{\bfomega \in \Sigma_k\\ \bfeta \in \Gamma^k}} \, \inf_{\bfpi \in \Pi(\bfupsilon)} -\sum_{i = 1}^n \sum_{j = 1}^k \pi_{ij} \log(\omega_j p_{\eta_j}(x_i)) + \lambda H(\bfpi) \enspace.
\end{equation}
Thanks to this relaxation, we can avoid the computationally expensive resolution of the initial transport problem~\eqref{generalproblem} with respect to $\bfpi\in \Pi(\bfupsilon, \bfomega)$. As will be detailed in section~\ref{sec:optim}, the expectation step of our algorithm (i.e. the estimation of $\bfpi$) is rather a simple closed form projection on $\Pi(\nu)$. 
We now show that solutions of the relaxed problem~\eqref{problem}  provide solutions for problem~\eqref{generalproblem}. 
To that end, we first introduce the notion of coordinatewise minimum~\cite{tseng2001convergence}. We say that $x^*\in\R^n$ is a coordinatewise minimum  of the nonconvex problem $\min_x f(x_1,x_2,\cdots x_n)$ if for all $i=1\cdots n$, $x^*_i=\uargmin{t}f(x^*_1,x^*_{i-1},t,x^*_{i+1},\cdots x^*_n)$. The next proposition states that local minima of problem~\eqref{problem} are local minima of problem~\eqref{generalproblem}.
\begin{proposition}\label{prop1} Any  coordinatewise minimum   of the relaxed problem~\eqref{problem} $(\hat\bfpi, \hat\bfomega,\hat\bfeta)$ is an admissible solution and a coordinatewise minimum of  problem~\eqref{generalproblem}.
\end{proposition}
The proof of this proposition is given in the section~\ref{sec:prop1} of the appendix. It first relies on the fact that the minimal value of the objective function in the relaxed problem is smaller than the one of the constrained problem~\eqref{generalproblem}. The proof then consists in showing that a minimizer of the relaxed problem necessarily belongs to set of constraints of the original problem~\eqref{generalproblem}.

\section{$\lambda-$EM algorithm}\label{sec:optim}
We propose to solve problem~\eqref{problem} using  block coordinate descent. This gives  an alternate optimization scheme on the transport plan $\bfpi$, mixture weights $\bfomega$ and mixture parameters $\bfeta$. 
We present the optimization steps in section~\ref{ssec:alt_opt}. In section~\ref{ssec:conv}, we analyze the convergence properties of the algorithm. 
In section~\ref{ssec:link} we  detail the connection of our framework with hard and soft clustering algorithms.  
\subsection{Alternate Optimization}\label{ssec:alt_opt}

Problem \eqref{problem} being nonconvex in $(\bfpi, \bfomega,\bfeta)$, we propose to consider an alternate procedure to estimate a local minima.
The corresponding algorithm consists in iteratively fixing two variables and optimizing with respect to the last one, following the update order $(\bfpi, \bfomega,\bfeta)$. We now describe each optimization step.

\paragraph{Transport plan.} For fixed weights $\bfomega \in \Sigma_k$ and parameters $\bfeta \in \Gamma^k$, the transport plan $\bfpi \in \Pi(\bfupsilon)$ is updated by solving a relaxed ROT  problem:
\begin{equation}
\label{pi}
\min_{\bfpi \in \Pi(\bfupsilon)} -\sum_{i = 1}^n \sum_{j = 1}^k \pi_{ij} %\upsilon_i 
\log(\omega_j p_{\eta_j}(x_i)) + \lambda H(\bfpi) \enspace.
\end{equation}

Recalling that the entropic regularization is defined as  $H(\bfpi)=\sum_{ij}\pi_{ij}(\log(\pi_{ij})-1)$, 
the global optimum of \eqref{pi} for fixed $\bfeta$ and $\bfomega$ writes:\vspace{-0.1cm}
\begin{equation}\label{expect}
\pi_{ij}^* = \upsilon_i\frac{(\omega_j p_{\eta_j}(x_i))^{1 / \lambda}}{ \sum_{l = 1}^k (\omega_l p_{\bfeta_l}(x_i))^{1 / \lambda}} \enspace.
\end{equation}
As discussed in Appendix~\ref{app:gen}, this step can be generalized to other stringly convex regularizers $H$.

\paragraph{Weights.} For fixed parameters $\bfeta \in \Gamma^k$ and transport plan $\bfpi \in \Pi(\bfupsilon)$, we are now left out with the maximization of a concave objective: 
%\begin{equation}
% \label{omega}
$\max_{\bfomega \in \Sigma_k} \sum_{i = 1}^n \sum_{j = 1}^k \pi_{ij} \log(\omega_j)$. % \enspace.
%\end{equation}
The solution is actually reached at $\bfomega = \bfpi^\top \ones_n$, i.e. 
\begin{equation}
\label{omega2}\omega_j = \sum_{i = 1}^n \pi_{ij}.\end{equation}
If   a cluster is empty after a weight update, it  is  removed for the remaining steps of the algorithm.

\paragraph{Parameters.} For fixed weights $\bfomega \in \Sigma_k$ and transport plan $\bfpi \in \Pi(\bfupsilon)$, if weights are non zero, the updates of parameters $\bfeta \in \Gamma^k$ can be solved independently for each cluster:
\begin{equation}
\label{xi}
\sup_{\eta_j \in \Gamma} \sum_{i = 1}^n \pi_{ij} \log(p_{\eta_j}(x_i)) \enspace.
\end{equation}
This step is similar to maximum likelihood estimation of the parameters, since it aims to maximize a weighted log-likelihood. In Sec.~\ref{sec:exp_fam}, we specify the update for the exponential family mixture model.

\subsection{Convergence analysis}\label{ssec:conv}

As the process alternates between minimization steps, the sequence of iterates may admit accumulation points. Using results from~\cite{grippo2000convergence}, we  show in the next theorem that such accumulation points are necessarily stationary points of the  nonconvex problem~\eqref{problem}. 
\begin{definition} Let $f(x_1, ...,x_n)$ be a differentiable  function defined that {\bf\em may not be convex}. A point $\hat x = (\hat x_1,\ldots,\hat x_n)$ is a stationary point iff $\forall y$, $\langle \nabla f(\hat x), y - \hat x \rangle \geq 0$. 
\end{definition}

\begin{theorem}[Proof in section~\ref{ssec:convergence} of the Appendix]\label{thm} Let $(\hat\bfpi, \hat\bfomega,\hat\bfeta)$ be an accumulation point of the algorithm given by the three steps~\eqref{expect},~\eqref{omega2} and~\eqref{xi}. If (i) $\exists B \subset \Gamma^k$ closed and convex so that all values resulting from step~\eqref{xi} satisfy $\bfeta \in B$, (ii) $\forall (i,j), \enspace \log(p_{\eta_j}(x_i))$ is continuously differentiable along $\eta_j$ in the neighborhood of $\hat \eta$, and (iii) $\hat \pi_{ij} > 0$, $\hat \omega_j > 0$, then $(\hat\bfpi, \hat\bfomega,\hat\bfeta)$ is a stationary point of the problem~\eqref{problem} in $\Pi(\upsilon) \times \Sigma_k \times B$. Moreover, if $\forall i,j \enspace -\log(p_{\eta_j}(x_i))$ is convex along $\bfeta$, $(\hat\bfpi,\hat\bfomega,\hat\bfeta)$ is a coordinatewise minimum of the function.

\end{theorem}

In the case of GMMs, that belong to the exponential family (see section \ref{ssec:gmm}), the  parameter set $\Gamma$ for a Gaussian is $\{ \eta \in \R^d\times \R^{d\times d}, \exists \nu, \Sigma \in S_d^{++}, \eta = [\Sigma^{-1}\nu, -\frac{1}{2}\Sigma^{-1}]\}$. We can define a minimum threshold $\epsilon>0$ so that the minimum eigenvalue of variance matrix $\Sigma$ is greater to $\epsilon$ and thus make  $\Gamma$ a closed convex set $\{ \eta \in \R^d\times \R^{d\times d}, \exists \nu, \Sigma -\epsilon I \in S_d^{+}, \eta = [\Sigma^{-1}\nu, -\frac{1}{2}\Sigma^{-1}]\}$ in which $A(\eta)$ is continuously differentiable. Finally, as for all exponential families, $-\log(p_{\eta_j}(x_i)) = -\log(h(x_i)) -\langle \eta_j, T(x_i) \rangle + A(\eta_j)$ is convex. Therefore, for GMMs any accumulation point is a coordinatewise minimum of the function.

We finally detail the evolution of the log-likelihood  $\log(p(\bfx|\bfomega,\bfeta))$ during our generalized algorithm. This result show that our algorithms aims at estimating a parametric model that  maximizes the log-likelihood of the data $x$.
\begin{proposition}[Proof in section~\ref{ssec:likelihood} of the appendix] \label{prop2}
At each step~\eqref{expect},~\eqref{omega2} and~\eqref{xi}, the function $\log(p(\bfx|\bfomega,\bfeta))+\lambda\log(||p^{1/\lambda}(\bfpi|\bfx,\bfomega,\bfeta||_1)$ is non decreasing.
When $\lambda \leq 1$, the log-likelihood $\log(p(\bfx|\bfomega,\bfeta))$ is therefore bounded from below  during the algorithm by a non decreasing function. 
\end{proposition}

\subsection{Hard and soft clustering; and beyond}\label{ssec:link}
Depending on the choice of the regularization parameter, the proposed framework interpolates between hard and soft clustering for $\lambda\in[0,1]$, and it recovers the maximum likelihood estimator (MLE) for $\lambda\to+\infty$.
First notice that for $\lambda=0$, the transport plan optimization~\eqref{pi} simplifies as 
\begin{equation}
\label{optim:pi}
\inf_{\bfpi \in \Pi(\bfupsilon)} -\sum_{i = 1}^n \sum_{j = 1}^k \pi_{ij} %\upsilon_i
 \log(\omega_j p_{\eta_j}(x_i)) \enspace.
\end{equation}
In this setting, the constraint $\bfpi \in \Pi(\bfupsilon)$ allows to solve the problem separately for each point $x_i$: if the value function $\gamma_{ij}=-\log(\omega_j p_{\eta_j}(x_i))$ is higher for a cluster $j=j^*$ compared to the others, the solution is attributing the point's mass $v_i$ to it: $\pi_{ij^*} = v_i$ and $\pi_{ij} = 0$ for $j \neq j^*$. The algorithm is thus a hard clustering algorithm. We illustrate this property in Appendix~\ref{app:hard}.

In the  case $\lambda=1$, the transport plan~\eqref{expect} corresponds to the "a posteriori" distribution $p(\bfpi|\bfx,\bfomega,\bfeta)$ of the likelihood of  a realization to be issued from the different clusters.   
This is a soft clustering of data points $x_i$ with respect to all components $j$,  that exactly corresponds to the EM algorithm.

When $\lambda = +\infty$, problem~\eqref{pi} is similar to a maximum likelihood: the entropy is minimum for  $\omega_j=1/k$ so that all components have the same contribution within the obtained parameterization.

\section{Application to Exponential Family Mixture Models}\label{sec:exp_fam}
In this section, we apply our algorithm to Exponential Family mixtures models. We show in section~\ref{ssec:exp_fam} that in this setting, all optimization steps admit closed form solutions. We explicitly write out the computation for the case of Gaussian Mixture Models (GMM) in section~\ref{ssec:gmm}. 

\subsection{Parameter Estimation for Exponential Family Components}\label{ssec:exp_fam}
%\subsection{Formulation}
We consider a mixture of minimal exponential family components with natural parameters $\bfeta$~\cite{mjordanexpf}:
\begin{equation}\label{eq:natural}
p_{\bfeta}(\bfx)=h(\bfx)\exp(T(\bfx)^\top \bfeta - A(\bfeta)) \enspace,
\end{equation}
where $T$ is a transformation on $\bfx$  and  the cumulant function $A$ is strictly convex and differentiable. Since $h$ only depends on the variable $\bfx$, we can remove this term in the following. Moreover, by introducing ${\bfmu}= \E[T(X)]$, there is a bijective mapping between ${\bfmu}$ and $\bfeta$ through ${\bfmu} = \nabla A (\bfeta)$. Therefore, the law $p_{\bfeta}$ can be parameterized along ${\bfmu}$.
In this setting, step~\eqref{xi} writes:
\begin{equation}\label{pb_efm}
\sup_{\eta_j} \sum_{i=1}^n \pi_{ij} (T(x_i)^\top \eta_j - A(\eta_j)) .
\end{equation}

As  parameters update  follows weights update, empty clusters have been removed and we have $\omega_j = \sum_{i = 1}^n \pi_{ij}>0$. The parameter optimization step  thus corresponds to solving the problem
$\sup_{\eta_j}  \sum_{i=1}^n \frac{\pi_{ij}}{\omega_j}T(x_i)^\top \eta_j - A(\eta_j)$, 
that involves a strictly concave function whose  gradient vanishes for
%\begin{equation}\label{step_efm} 
$\mu_j=\nabla A(\eta_j) = \sum_{i=1}^n \frac{\pi_{ij}}{\omega_j}T(x_i)$. %\end{equation}
Then, if $\hat \mu_j = \sum_{i=1}^n \frac{\pi_{ij}}{\omega_j}T(x_i) $ belongs to the definition set of the family barycentric parameter, it solves the parameter optimization step~\eqref{pb_efm}.

The overall inference process for exponential families is given in Algorithm~\ref{algo}.
\begin{algorithm}\label{algo}
\caption{Parameter estimation algorithm for Exponential Family Mixture Models}
\label{algorithm}
\begin{algorithmic}
\STATE Initialize $\bfomega$ and $\bfmu$ 
\REPEAT
\STATE  Expectation: $\pi_{ij}^* = \upsilon_i\frac{(\omega_j p_{\mu_j}(x_i))^{1 / \lambda}}{\sum_{l = 1}^k (\omega_l p_{\mu_l}(x_i))^{1 / \lambda}}$
\STATE  Weight update: $\omega_j = \sum_{i = 1}^n \pi_{ij}$
\STATE Maximization:  $\mu_j =\sum_{i=1}^n\frac{\pi_{ij}}{\omega_j}T(x_i)$
\UNTIL{convergence}
\end{algorithmic}
\end{algorithm}

\subsection{Case of Gaussian Mixture Models}\label{ssec:gmm} Gaussian Mixture Model (GMM) is the most well-know model of the exponential family.  For a $1D$ Gaussian of mean value $\nu$ and variance $\sigma^2$, expression~\eqref{eq:natural} with natural parameters is obtained with $$\eta = (\frac{\nu}{\sigma^2},\frac{-1}{2\sigma^2})^T,\;\; T(x) = (x, x^2)^T, \;\;h(x) = (2\pi)^{-1/2} \textrm{ and }A(\eta) = -\frac{\eta_1^2}{4 \eta_2} - \frac{1}{2}\log( -2 \eta_2),$$ so as to recover 
%\begin{align*}
    $p_{\eta}(x)    = h(x)\exp(\eta_1 x +\eta_2 x^2 - (-\frac{\eta_1^2}{4 \eta_2} - \frac{1}{2}\log( -2 \eta_2)) 
     = \frac{1}{\sqrt{2\pi}\sigma}\exp(-\frac{(x-\nu)^2}{2\sigma^2})$.  %\enspace .
%\end{align*}

%In our numerical experiments, we  focus on the inference of 2D GMMs.
In higher dimensions, multivariate normal law parameterized with mean $\nu$ and variance $\Sigma$ remain within the exponential family by taking:
$$ \eta = (\Sigma^{-1}\nu, -\frac{1}{2} \Sigma^{-1})^T,\;\;T(x) = (x, xx^T)^T \textrm{ and } A(\eta) = -\frac{1}{4}\eta_1^T\eta_2^{-1}\eta_1 - \frac{1}{2}\log|-2\eta_2|.$$

After the maximization step~\eqref{pb_efm}, the update of the parameters  $(\nu_j,\Sigma_j)$ of the GMM components $p_{\eta_j}$  are obtained at the expectation step using $\mu_j =\sum_{i=1}^n\frac{\pi_{ij}}{\omega_j}T(x_i) = [\nu_j, \Sigma_j + \nu_j \nu_j^T]$, so that  $\nu_j = (\mu_j)_1$ and $ \Sigma_j= (\mu_j)_2 - (\mu_j)_1 (\mu_j)_1^T $.%\mu = \E[T(x)] = [\nu, \Sigma + \nu \nu^T]$. % hence $\nu = \mu_1$ and $\Sigma = [\mu_2 - \mu_1 \mu_1^T ]$. 

\section{Experiments}\label{sec:expe}
Our experiments focus on 1D and 2D Gaussian Mixture Models (GMMs) for the reference model, to be able to  measure the quality of the inference. The  objective is to analyze the effect of the regularization parameter $\lambda$ on the inferred GMM.

\subsection{Illustration with inference of 1D GMM}

\begin{figure}[h]
\begin{center}
    \begin{tabular}{cc}
      \includegraphics[width=.45\textwidth,  height=2.2cm,trim={2cm 0cm 0cm 0cm},clip]{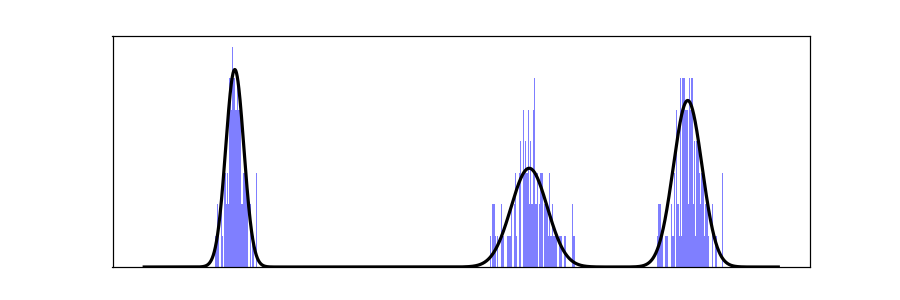}\vspace{-0.15cm}&\includegraphics[width=.45\textwidth,  height=2.2cm,trim={2cm 0cm 0cm 0cm},clip]{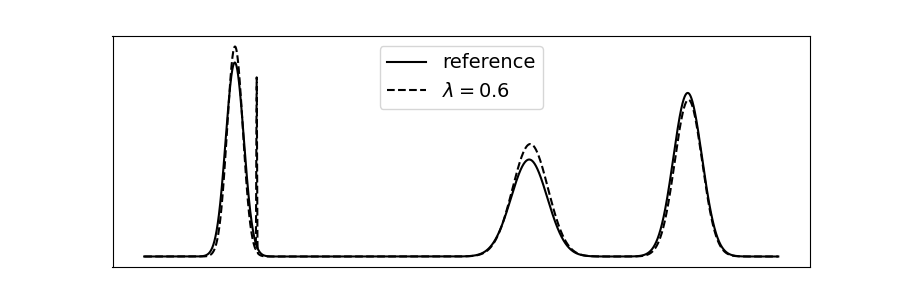}\vspace{-0.15cm}\\
      (a) Histogram of samples (in blue)&
    (b) Inference with $\lambda=0.6$\\
    from the reference distribution (black line)& (error $W_2=5.7$)\\
     \includegraphics[width=.45\textwidth,  height=2.2cm,trim={2cm 0cm 0cm 0cm},clip]{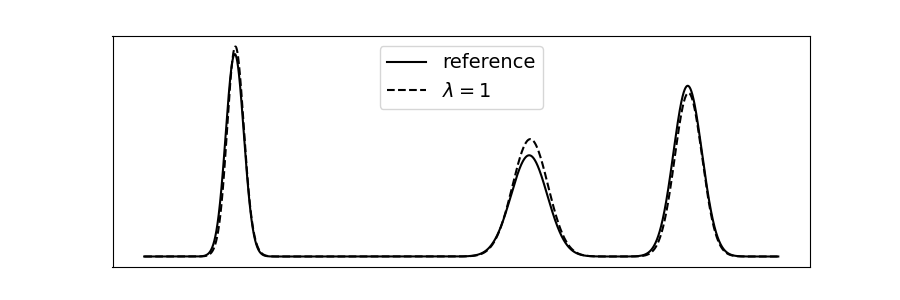}&    \includegraphics[width=.45\textwidth,  height=2.2cm,trim={2cm 0cm 0cm 0cm},clip]{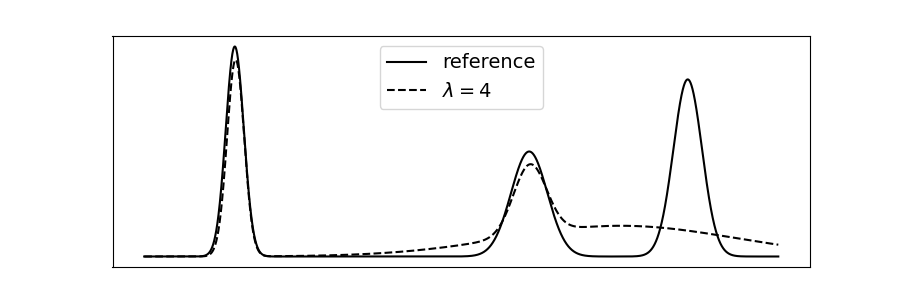}\vspace{-0.15cm}\\
     (c) Inference with $\lambda=1$ &
     (d) Inference with $\lambda=4$ \\
     (error $W_2=5.7$)&(error $W_2=11.5$)\vspace*{-0.2cm}
         \end{tabular}
    \caption{\label{fig:inferences_1D}Examples of inference of a 1D GMM.\vspace*{-0.2cm}   %(a)  and  inference results for (b) $\lambda=0.6$ , (c) $\lambda=1$ and (d) $\lambda=4$.
    }
    \end{center}
\end{figure}

We infer a 1D Gaussian Mixture Model on samples  issued from a reference GMM model (see figure~\ref{fig:inferences_1D}(a)). To validate our model we consider the Wasserstein distance between the discretization of our reference model and the inferred ones. In 1D, the Wasserstein distance indeed admits a closed form expression  $W_2(\alpha, \beta) = ( \int_{0}^{1} |Q_\alpha(x) - Q_\beta(x)|^2 \, \mathrm{d}x)^{\frac{1}{2}}$; where $\alpha$  and $\beta$ are measures and $Q_\alpha$ is the quantile function associated to $\alpha$. 

To illustrate the effect of the regularization, we considered a  GMM with $4$ components for inference, while the reference GMM only contains $3$ components. A k-means algorithm is used to initialize the clusters. As shown in figure~\ref{fig:inferences_1D}(b-d), when $\lambda = 1$, the inferred distribution is close to the reference one ({$W_2=5.7$}). For a smaller value $\lambda = 0.6$, the wasserstein distance remains almost identical ({$W_2=5.7$}), but we observe overfitting, as  one inferred cluster gets locally stuck on a particular data point.  This illustrates that a parameter $\lambda<1$ may lead to an inferred model that is not robust to data outliers, i.e. that can not escape from a cluster $k$ currently associated to a single isolated data point. Finally, for $\lambda = 4$, the inferred GMM is less accurate ({$W_2=11.4$}), as it contains a wider cluster mixing two clusters from the distribution of the reference model. On the other hand, taking values  $\lambda>>1$ makes the inferred distribution spread out,  thus failing at localizing accurately the reference clusters.

\subsection{Quantitative evaluation on 2D GMMs}

%\subsubsection{Experimental setting}
In order to quantify the effect of the regularization parameter on inference quality, we compare the reference and inferred models with the $MW_2$ distance introduced in~\cite{delon2020wassersteintype}, that provides an upper bound  of the true Wasserstein Distance between GMMs. The $MW_2$ distance  is obtained by solving a simple OT problem  whose dimension only depends on the number of clusters in the inferred and reference models.

\begin{wrapfigure}[12]{r}{7.5cm}
    \centering \vspace*{-0.4cm}
    \hspace{-0.4cm}\includegraphics[width=.45\textwidth,trim={1.3cm 0cm 2cm 1cm},clip]{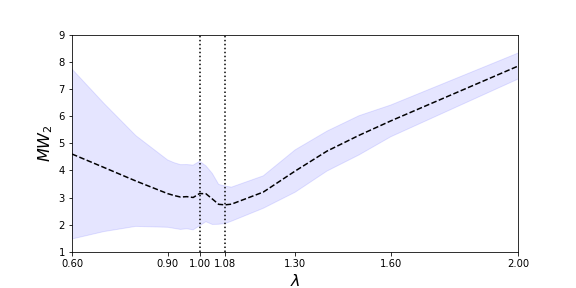}\vspace{-0.1cm}
    \caption{\label{fig:variabilite_inferences_2D}Mean and variance of the inference accuracy ($MW_2$ distance), for various regularization levels $\lambda$.}
\end{wrapfigure}

\paragraph{Influence of $\lambda$ and robustness to initialization}
For  $20$ random reference GMMs with 3 clusters, we sample $n=1000$ points and consider the  inference of GMMs with $k=5$ components for different $\lambda$ and $20$ different random initializations for each experiment.
As shown in Figure~\ref{fig:variabilite_inferences_2D}, best accuracy (in terms of $MW_2$ distance) is reached for $\lambda\approx 1.08$, whereas  performances deteriorate 
  for $\lambda<0.9$ or  $\lambda>1.2$. All in all, taking $\lambda = 1.1$ improves both robustness (smaller standard deviation of the $MW_2$ distance) and accuracy compared to the EM algorithm $\lambda=1$.

\begin{figure}[h]
\begin{centering}
    \begin{tabular}{ccc}
    \includegraphics[width=.25\textwidth,trim={2.2cm 3cm 2.6cm 2.8cm},clip]{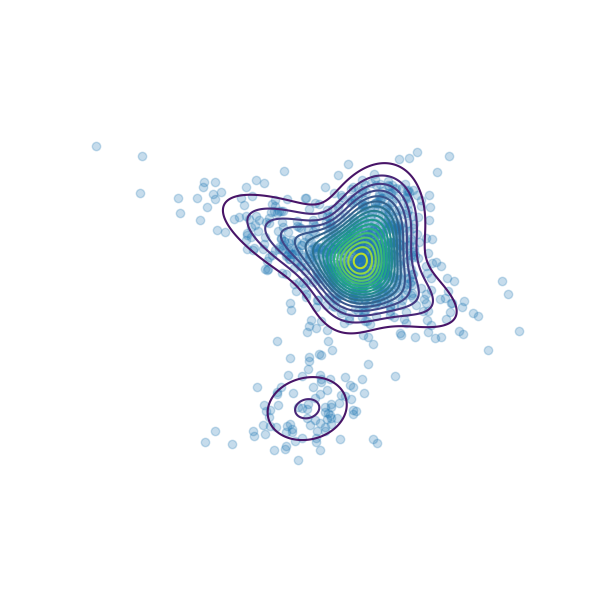}&\includegraphics[width=.25\textwidth,trim={2.2cm 3cm 2.6cm 2.8cm},clip]{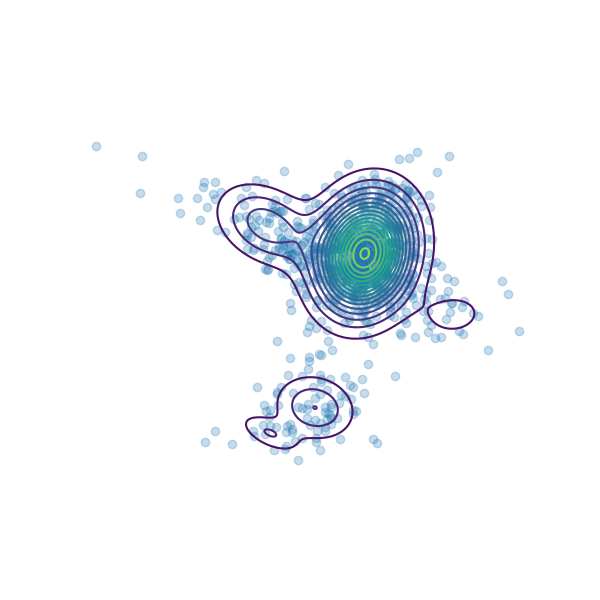}&  \includegraphics[width=.25\textwidth,trim={2.2cm 3cm 2.6cm 2.8cm},clip]{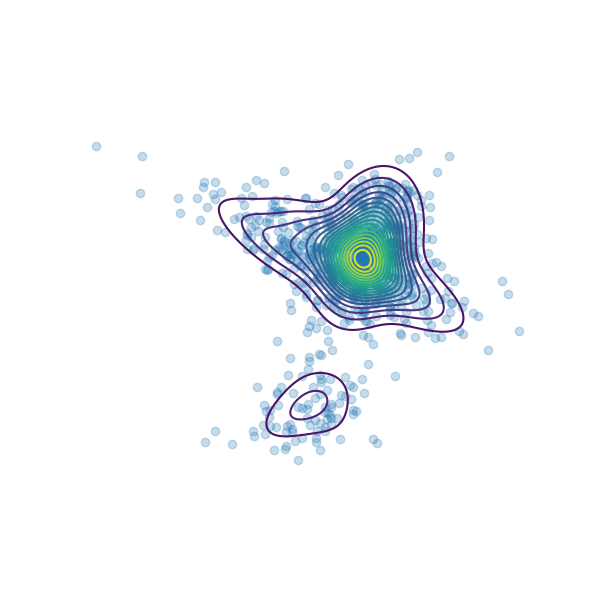}\\
    (a) Reference&(b) $\lambda=0.6$&(c) $\lambda=1$\\& (error $MW_2 = 3.7$)&(error $MW_2 = 2.0$)\\
        \includegraphics[width=.25\textwidth,trim={2.2cm 3cm 2.6cm 2.8cm},clip]{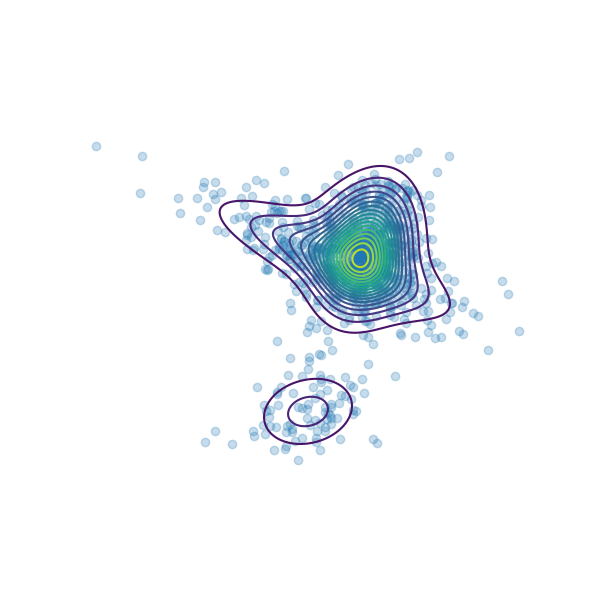}&
        \includegraphics[width=.25\textwidth,trim={2.2cm 3cm 2.6cm 2.8cm},clip]{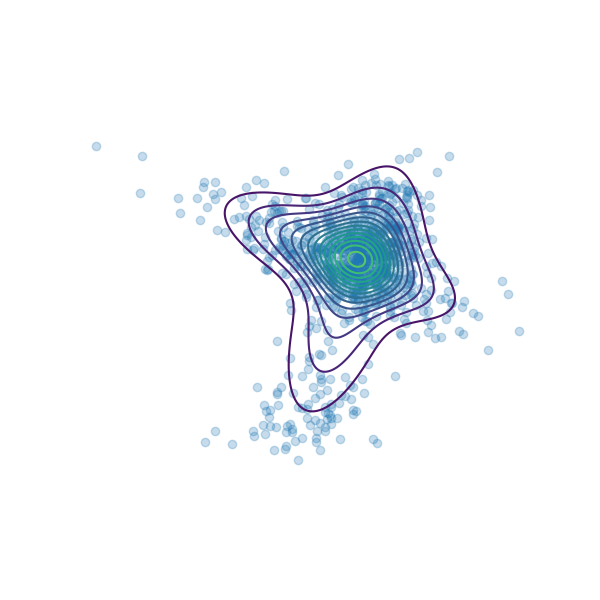}&
    \includegraphics[width=.25\textwidth,trim={2.2cm 3cm 2.6cm 2.8cm},clip]{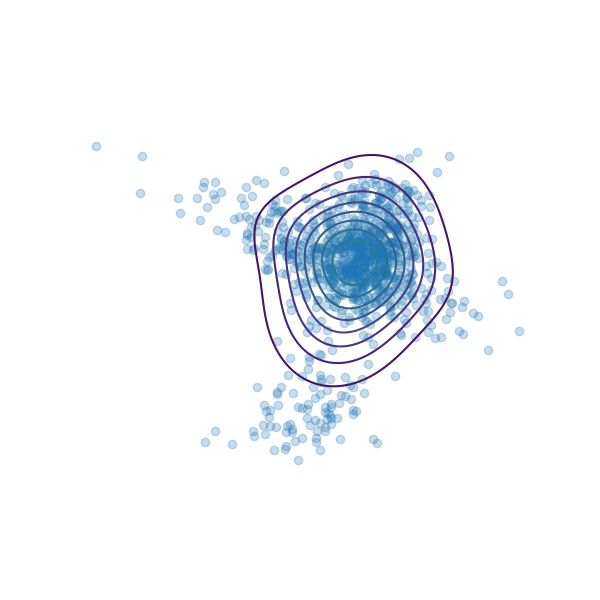}\\
    (d) $\lambda=1.1$&(e) $\lambda=1.6$&(f) $\lambda=3$\\ (error $MW_2 = 1.3$)&(error $MW_2 = 3.9$)& (error $MW_2 = 5.1$)
    \end{tabular}
    \caption{\label{fig:inferences_2D}2D Inferences for several values of $\lambda$ }
\end{centering}
\end{figure}

We provide in  Figure~\ref{fig:inferences_2D} a qualitative illustration. As observed for $1D$ GMMs, for smaller values of $\lambda$, clusters may concentrate on small group of points, at the expense of the global shape recovery of the reference GMM. Inference performance is increased in  the setting  $\lambda = 1$, which corresponds to the EM algorithm. Finally $\lambda = 1.1$ is the optimal value for the inference. It provides a more regular shape, closer to the reference distribution. With higher values $\lambda>1.5$, the inferred distribution becomes close to the maximum likelihood estimator with only one cluster.

\begin{figure}[h]
    \centering
       \begin{tabular}{cc}\includegraphics[width=.38\textwidth,trim={1.3cm 0cm 2cm 1cm},clip]{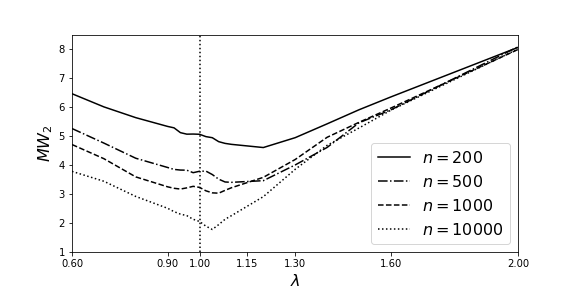}&\includegraphics[width=.38\textwidth,trim={1.3cm 0cm 2cm 1cm},clip]{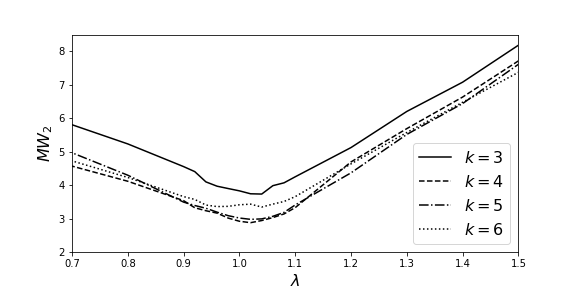}\\
    (a)&(b)
       \end{tabular}
    \caption{\label{fig:influence_nbpoints}(a) Inference accuracy, in terms of  $MW_2$ distance, for various number of samples $n$ and different regularization levels $\lambda$. (b) Inference accuracy, in terms of  $MW_2$ distance, for various number of clusters $k$ in the inferred GMMs, and different regularization levels $\lambda$.}
\end{figure}

\paragraph{Number of sampled points}

Here we consider the average inference performance over 80 reference GMMs with $k=4$ clusters and a KM initialization, and we study the influence of the number of sampled points on the optimal $\lambda$.
Figure~\ref{fig:influence_nbpoints}(a) highlights that, for a lower (resp. larger) number of samples the optimal regularization value is close to $\lambda = 1.2$  (resp. $\lambda = 1.05$). We suggest that  that for a lower number of points, higher value of $\lambda$ leads to a more robust inference. 

\paragraph{Number of clusters}

We finally consider the influence of the number of clusters in the inferred distribution in order to study both underfitting and overfitting behaviour with respect to $\lambda$. We consider $80$ random reference GMMs with $k=4$ clusters. 
As shown in Figure~\ref{fig:influence_nbpoints}(b), the best inference is obtained with  $\lambda=1.02$ and the true number of cluster $k=4$. A small  decrease of performance is observed for $k=5$. In all cases, a parameter $\lambda\approx 1.05 $ allows to be robust to the uncertainty of the number of clusters for the inferred GMM.

\subsection{Classification in higher dimension}
\begin{figure}[h]
    \centering
\includegraphics[width=.5\linewidth,height=.4\linewidth]{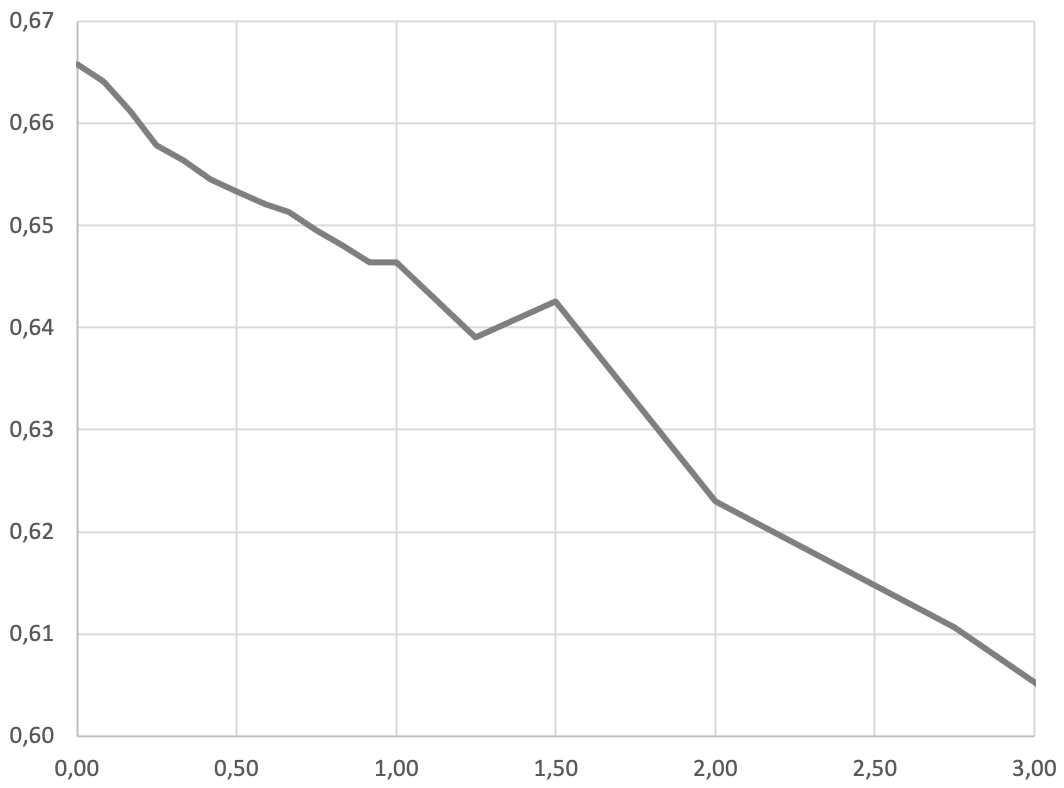}
\caption{\label{HC2}Classification accuracy of the MNIST database for $\lambda\in[0,3]$.}
\end{figure}

\begin{figure}[h]
    \centering
\includegraphics[width=.8\linewidth,height=.18\linewidth]{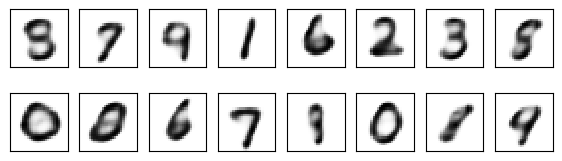}
\caption{\label{HC3}Centroids of the $k=16$ clusters obtained by  our EM method with $\lambda=10^{-3}$ on the latent space ($n=64$) of an autoencoder learnt on MNIST.}
\end{figure}

We finally present an experiment in higher dimension to illustrate the interest of the proposed method for classification. We realize a clustering in $k=16$ groups of $2000$ samples of the latent space (dimension $n=64$) of an autoencoder learnt on MNIST dataset. We display in Figure~\ref{HC2} the accuracy of the corresponding clustering for $\lambda=0$ to $\lambda=2$. A better accuracy ($0.666)$ is obtained with $\lambda\to 0.$ than with the classical EM (accuracy $0.647$, almost identical to the  scikit-learn implementation of  EM that gives an accuracy of $0.645$) reached with $\lambda=1$  or higher values (accuracy 0.605 for $\lambda=3$).
This suggests that choosing hard clustering with $\lambda=0$ is a relevant choice in term of accuracy for clustering purpose.
We display in Figure~\ref{HC3} the centroids of the computed clusters for $\lambda=10^{-3}$.

\section{Conclusion}
In this paper we reformulate the parameter estimation problem in finite mixture models from the point of view of regularized optimal transport. Considering Sinkhorn distance, we are able to recover standard algorithms such as Maximum Likelihood, EM or $k-$means  as specific instances of our general $\lambda$-EM algorithm. 
Our numerical results highlight the interest of taking a parameter $\lambda\approx 1.1$ to increase the robustness of the classical EM to initialization and to few data points, while considering hard clustering with $\lambda\to 0$ for classification purposes.

Future works may include hierarchical clustering, other ROT models~\cite{dessein2018regularized} or costs penalizing data likelihood, ground metric learning~\cite{Cuturi2014}, total Bregman divergences~\cite{Liu2012}, relaxed data point weights~\cite{wang2017robust}  through unbalanced optimal transport or  annealing strategies to avoid poor local minima~\cite{mandt2016variational}.

\paragraph{Acknowledgements} This study has been carried out with financial support from the French State, managed by the French National Research Agency (ANR) in the frame of the ``Investments for the future'' Program IdEx Bordeaux (ANR-10-IDEX-03-02), Cluster of excellence CPU and the GOTMI project (ANR-16-CE33-0010-01).

\bibliographystyle{splncs04}

\begin{thebibliography}{10}
\providecommand{\url}[1]{\texttt{#1}}
\providecommand{\urlprefix}{URL }
\providecommand{\doi}[1]{https://doi.org/#1}

\bibitem{adler2018banach}
Adler, J., Lunz, S.: Banach wasserstein gan. Adv. in Neur. Inf. Proces. Syst.
  \textbf{31} (2018)

\bibitem{Carlier_wasserstein_barycenter}
Agueh, M., Carlier, G.: Barycenters in the wasserstein space. SIAM Journal on
  Mathematical Analysis  \textbf{43}(2),  904--924 (2011)

\bibitem{arjovsky2017wasserstein}
Arjovsky, M., Chintala, S., Bottou, L.: Wasserstein generative adversarial
  networks. In: Int. Conf. on Machine Learning. pp. 214--223. PMLR (2017)

\bibitem{Arthur2007}
Arthur, D., Vassilvitskii, S.: k-means++: The advantages of careful seeding.
  In: CM-SIAM Symposium on Discrete Algorithms. pp. 1027--1035 (2007)

\bibitem{Banerjee2005}
Banerjee, A., Merugu, S., Dhillon, I.S., Ghosh, J.: Clustering with {Bregman}
  divergences. Journal of Machine Learning Res.  \textbf{6},  1705--1749 (Dec
  2005)

\bibitem{del2016robust}
del Barrio, E., Cuesta-Albertos, J.A., Matr{\'a}n, C., Mayo-{\'I}scar, A.:
  Robust k-barycenters in wasserstein space and wide consensus clustering.
  arXiv preprint arXiv:1607.01179  (2016)

\bibitem{BASSETTI20061298}
Bassetti, F., Bodini, A., Regazzini, E.: On minimum kantorovich distance
  estimators. Statistics \& Probability Letters  \textbf{76}(12),  1298 -- 1302
  (2006)

\bibitem{Basu1998}
Basu, A., Harris, I.R., Hjort, N.L., Jones, M.C.: Robust and efficient
  estimation by minimising a density power divergence. Biometrika
  \textbf{85}(3),  549--559 (Sep 1998)

\bibitem{Basu2011}
Basu, A., Shioya, H., Park, C.: Statistical Inference: The Minimum Distance
  Approach, Monographs on Statistics and Applied Probability, vol.~120 (2011)

\bibitem{Benaroya2006}
Benaroya, L., Bimbot, F., Gribonval, R.: Audio source separation with a single
  sensor. IEEE Trans. on Audio, Speech, and Language Proces.  \textbf{14}(1),
  191--199 (2006)

\bibitem{bernton2019parameter}
Bernton, E., Jacob, P.E., Gerber, M., Robert, C.P.: On parameter estimation
  with the wasserstein distance. Information and Inference: A Journal of the
  IMA  \textbf{8}(4),  657--676 (2019)

\bibitem{blanchet2021sample}
Blanchet, J., Kang, Y.: Sample out-of-sample inference based on wasserstein
  distance. Operations Research  \textbf{69}(3),  985--1013 (2021)

\bibitem{bousquet2017optimal}
Bousquet, O., Gelly, S., Tolstikhin, I., Simon-Gabriel, C.J., Schoelkopf, B.:
  From optimal transport to generative modeling: the vegan cookbook. arXiv
  preprint arXiv:1705.07642  (2017)

\bibitem{Carson2002}
Carson, C., Belongie, S., Greenspan, H., Malik, J.: Blobworld: Image
  segmentation using expectation-maximization and its application to image
  querying. IEEE Trans. Pat. Analys. and Mach. Intel.  \textbf{24}(8),
  1026--1038 (2002)

\bibitem{chen2018optimal}
Chen, Y., Georgiou, T.T., Tannenbaum, A.: Optimal transport for gaussian
  mixture models. IEEE Access  \textbf{7},  6269--6278 (2018)

\bibitem{cuturi2013sinkhorn}
Cuturi, M.: Sinkhorn distances: Lightspeed computation of optimal transport.
  Adv. in Neural inf. proces. syst.  \textbf{26} (2013)

\bibitem{Cuturi2014}
Cuturi, M., Avis, D.: Ground metric learning. Journal of Machine Learning Res.
  \textbf{15}(1),  533--564 (2014)

\bibitem{delon2020wassersteintype}
Delon, J., Desolneux, A.: A wasserstein-type distance in the space of gaussian
  mixture models. SIAM Journal on Imaging Sciences  \textbf{13}(2),  936--970
  (2020)

\bibitem{dempster1977maximum}
Dempster, A.P., Laird, N.M., Rubin, D.B.: Maximum likelihood from incomplete
  data via the em algorithm. J. of the royal statistical soc. series B
  \textbf{39}(1),  1--22 (1977)

\bibitem{dessein2018regularized}
Dessein, A., Papadakis, N., Rouas, J.L.: Regularized optimal transport and the
  rot mover's distance. Journal of Machine Learning Res.  \textbf{19}(1),
  590--642 (2018)

\bibitem{Eguchi1983}
Eguchi, S.: Second order efficiency of minimum contrast estimators in a curved
  exponential family. The Annals of Statistics  \textbf{11}(3),  793--803
  (1983)

\bibitem{Eguchi2009}
Eguchi, S.: Information divergence geometry and the application to statistical
  machine learning. In: Inform. Theory and Stat. Learn., chap.~13, pp. 309--332
  (2009)

\bibitem{Eguchi2001}
Eguchi, S., Kano, Y.: Robustifying maximum likelihood estimation. Tech. rep.,
  The Institute of Statistical Mathematics (2001)

\bibitem{frogner2020approximate}
Frogner, C., Poggio, T.: Approximate inference with wasserstein gradient flows.
  In: Int. Conf. Artif. Intel. and Stat. pp. 2581--2590. PMLR (2020)

\bibitem{genevay2018learning}
Genevay, A., Peyr{\'e}, G., Cuturi, M.: Learning generative models with
  sinkhorn divergences. In: Int. Conf. Artif. Intel. and Stat. pp. 1608--1617.
  PMLR (2018)

\bibitem{greenspan2004probabilistic}
Greenspan, H., Goldberger, J., Mayer, A.: Probabilistic space-time video
  modeling via piecewise gmm. IEEE Trans. Pat. Analys. and Mach. Intel.
  \textbf{26}(3),  384--396 (2004)

\bibitem{grippo2000convergence}
Grippo, L., Sciandrone, M.: On the convergence of the block nonlinear
  gauss--seidel method under convex constraints. Operations res. let.
  \textbf{26}(3),  127--136 (2000)

\bibitem{ho2017multilevel}
Ho, N., Nguyen, X., Yurochkin, M., Bui, H.H., Huynh, V., Phung, D.: Multilevel
  clustering via wasserstein means. arXiv preprint arXiv:1706.03883  (2017)

\bibitem{houdard2022gradient}
Houdard, A., Leclaire, A., Papadakis, N., Rabin, J.: On the gradient formula
  for learning generative models with regularized optimal transport costs.
  Transactions on Machine Learning Research  (2022)

\bibitem{2016arXiv160500513I}
{Irpino}, A., {De Carvalho}, F., {Verde}, R.: {Fuzzy clustering of
  distribution-valued data using adaptive L2 Wasserstein distances}. ArXiv
  e-prints  (2016)

\bibitem{irpino2006new}
Irpino, A., Verde, R.: A new wasserstein based distance for the hierarchical
  clustering of histogram symbolic data. Data science and classification pp.
  185--192 (2006)

\bibitem{mjordanexpf}
Jordan, M.: Exponential families : basics. Berkeley course notes (2009)

\bibitem{Kearns1997}
Kearns, M., Mansour, Y., Ng, A.Y.: An information-theoretic analysis of hard
  and soft assignment methods for clustering. In: Conf. on Uncertainty in
  Artificial Intelligence. pp. 282--293 (1997)

\bibitem{lambert2022variational}
Lambert, M., Chewi, S., Bach, F., Bonnabel, S., Rigollet, P.: Variational
  inference via wasserstein gradient flows. Adv. in Neur. Inf. Proces. Syst.
  \textbf{35},  14434--14447 (2022)

\bibitem{Liu2012}
Liu, M., Vemuri, B.C., Amari, S.i., Nielsen, F.: Shape retrieval using
  hierarchical total {Bregman} soft clustering. IEEE Trans. Pat. Analys. and
  Mach. Intel.  \textbf{34}(12),  2407--2419 (2012)

\bibitem{Lloyd1982}
Lloyd, S.P.: Least square quantization in {PCM}. IEEE Trans. on Information
  Theory  \textbf{28}(2),  129--137 (1982)

\bibitem{mandt2016variational}
Mandt, S., McInerney, J., Abrol, F., Ranganath, R., Blei, D.: Variational
  tempering. In: Artificial intelligence and statistics. pp. 704--712. PMLR
  (2016)

\bibitem{marti2016optimal}
Marti, G., Andler, S., Nielsen, F., Donnat, P.: Optimal transport vs.
  fisher-rao distance between copulas for clustering multivariate time series.
  In: IEEE Statistical Signal Proces. Workshop. pp.~1--5 (2016)

\bibitem{mena2020sinkhorn}
Mena, G., Nejatbakhsh, A., Varol, E., Niles-Weed, J.: Sinkhorn em: an
  expectation-maximization algorithm based on entropic optimal transport. arXiv
  preprint arXiv:2006.16548  (2020)

\bibitem{Mihoko2002}
Mihoko, M., Eguchi, S.: Robust blind source separation by beta divergence.
  Neural Computation  \textbf{14}(8),  1859--1886 (2002)

\bibitem{NIPS2016_6248}
Montavon, G., M\"{u}ller, K.R., Cuturi, M.: Wasserstein training of restricted
  boltzmann machines. In: Adv. in Neur. Inf. Proces. Syst., pp. 3718--3726
  (2016)

\bibitem{Nielsen2012}
Nielsen, F.: {K-MLE}: A fast algorithm for learning statistical mixture models.
  In: IEEE Int. Conf. on Acoustics, Speech and Signal Proces. pp. 869--872
  (2012)

\bibitem{Nock2006}
Nock, R., Nielsen, F.: On weighting clustering. IEEE Trans. Pat. Analys. and
  Mach. Intel.  \textbf{28}(8),  1223--1235 (2006)

\bibitem{Pardo2006}
Pardo, L.: Statistical Inference Based on Divergence Measures. Statistics: A
  Series of Textbooks and Monographs, Chapman \& Hall/CRC (2006)

\bibitem{peyre2019computational}
Peyr{\'e}, G., Cuturi, M., et~al.: Computational optimal transport: With
  applications to data science. Foundations and Trends{\textregistered} in
  Machine Learning  \textbf{11}(5-6),  355--607 (2019)

\bibitem{Rubner2000}
Rubner, Y., Tomasi, C., Guibas, L.J.: The earth mover's distance as a metric
  for image retrieval. Int. Journal of Computer Vision  \textbf{40}(2),
  99--121 (2000)

\bibitem{Scheirer1997}
Scheirer, E., Slaney, M.: Construction and evaluation of a robust multifeature
  speech/music discriminator. In: IEEE Int. Conf. on Acoustics, Speech, and
  Signal Proces. vol.~2, pp. 1331--1334 (1997)

\bibitem{WKMEAN}
Staib, M., Jegelka, S.: Wasserstein k-means++ for cloud regime histogram
  clustering. In: Climate Informatics (2017)

\bibitem{tseng2001convergence}
Tseng, P.: Convergence of a block coordinate descent method for
  nondifferentiable minimization. Journal of optimization theory and
  applications  \textbf{109},  475--494 (2001)

\bibitem{Tzanetakis2002}
Tzanetakis, G., Cook, P.: Musical genre classification of audio signals. IEEE
  Trans. on Speech and Audio Processing  \textbf{10}(5),  293--302 (2002)

\bibitem{wang2017robust}
Wang, Y., Kucukelbir, A., Blei, D.M.: Robust probabilistic modeling with
  bayesian data reweighting. In: Int. Conf. on Machine Learning. pp.
  3646--3655. PMLR (2017)

\bibitem{ye2015fast}
Ye, J., Wu, P., Wang, J.Z., Li, J.: Fast discrete distribution clustering using
  {W}asserstein barycenter with sparse support. {IEEE} Trans. Signal Proces.
  \textbf{65}(9),  2317--2332 (2017)

\bibitem{yi2023sliced}
Yi, M., Liu, S.: Sliced wasserstein variational inference. In: Asian Conference
  on Machine Learning. pp. 1213--1228. PMLR (2023)

\bibitem{Zhang2009}
Zhang, J., Song, Y., Chen, G., Zhang, C.: On-line evolutionary exponential
  family mixture. In: Int. Joint Conf. on Artifical Intelligence. pp.
  1610--1615 (2009)

\bibitem{zhuang2022wasserstein}
Zhuang, Y., Chen, X., Yang, Y.: Wasserstein $ k $-means for clustering
  probability distributions. Adv. in Neur. Inf. Proces. Syst.  \textbf{35},
  11382--11395 (2022)

\bibitem{zoran2011learning}
Zoran, D., Weiss, Y.: From learning models of natural image patches to whole
  image restoration. In: IEEE Int. Conf. on Computer Vision. pp. 479--486
  (2011)

\end{thebibliography}

\appendix

\section{Generalized regularized optimal transport (ROT) models}\label{app:gen}
The model can be  extended to other ROT problems and linked with Bregman clustering algorithms~\cite{Banerjee2005}. Following~\cite{dessein2018regularized},  replacing the entropy $H$ by any differentiable and strongly convex potential $\phi$, the solution of problem~\eqref{pi} is equivalent to a Bregman projection:
$\min_{\bfpi \in \Pi(\bfupsilon)} B_\phi(\bfpi \Vert\nabla \phi^*( \log(\omega_j p_{\bfeta_j}(x_i)/\lambda)))$, 
where $\phi^*$ is the Legendre transform of $\phi$ and $
B_\phi(x \Vert y) = \phi(x) - \phi(y) - \langle x - y, \nabla\phi(y) \rangle $
 is the Bregman divergence associated to $\phi$.

The algorithm can also be efficiently implemented for any convex  regularizer $\phi(\bfpi)$ that is separable with respect to the first dimension of $\bfpi$ (i.e. $\phi(\bfpi)=\sum_{i}\phi(\bfpi_{i})$, with $\bfpi_i=\{\pi_{ij}\}_{j=1}^k$).
 As the constraint $\bfpi  \in \Pi(\bfupsilon)$ implies that $\bfpi_i\ones_k=v_i$, the problem can be solved in parallel for each $i$:
 \begin{equation}\label{optim:pidiff2}
\inf_{\bfpi_i\ones_k=v_i} -\sum_{j = 1}^k \pi_{ij} \log(\omega_j p_{\eta_j}(x_i))+\lambda \phi(\bfpi_{i})\enspace.
\end{equation}
\section{Constraint relaxation}\label{sec:prop1}

\textit{In this section, we present the detailed proof of Proposition~\ref{prop1}}

We first recall the general problem~\eqref{generalproblem}.

\begin{equation}
\label{generalproblem2}
\inf_{\substack{\bfomega \in \Sigma_k\\ \bfeta \in \Gamma^k}} \, \inf_{\bfpi \in \Pi(\bfupsilon, \bfomega)} -\sum_{i = 1}^n \sum_{j = 1}^k \pi_{ij}  %\upsilon_i 
\log(\omega_j p_{\eta_j}(x_i)) +  \lambda H(\bfpi) \;,
\end{equation}
with \begin{equation}
\Pi(\bfupsilon, \bfomega) = \{\bfpi \in \R_+^{n \times k} \colon \bfpi \ones_k = \bfupsilon, \bfpi^\top \ones_n = \bfomega\} \enspace;
\end{equation}
and the relaxed one~\eqref{problem}
\begin{equation}
\label{problem2}
\inf_{\substack{\bfomega \in \Sigma_k\\ \bfeta \in \Gamma^k}} \, \inf_{\bfpi \in \Pi(\bfupsilon)} -\sum_{i = 1}^n \sum_{j = 1}^k \pi_{ij} \log(\omega_j p_{\eta_j}(x_i)) + \lambda H(\bfpi) \enspace,
\end{equation}
for $\Pi(\bfupsilon) = \{\bfpi \in \R_+^{n \times k} \colon \bfpi \ones_k = \bfupsilon\}$.

\begin{proposition} If $(\hat\bfpi, \hat\bfomega,\hat\bfeta)$ is a  coordinatewise minimum of the relaxed problem \eqref{problem2} and if the objective function has a finite value at this point, then it is an admissible solution and a coordinatewise minimum of  problem \eqref{generalproblem2}.
\end{proposition}
\begin{proof}
First, for both problems, variables $\bfomega$ and $\bfeta$ are defined on the  same sets. Then, let $(\hat\bfpi, \hat\bfomega,\hat\bfeta)$ be a  coordinatewise minimum of the relaxed problem \eqref{problem2}. 
We introduce the Lagrangian function $\mathcal{L}$ corresponding to the minimisation subproblem in $\bfomega$ for fixed $\hat\bfpi$ and $\hat\bfeta$, with Lagrangian multiplier $\alpha$ for the constraint $\sum_j\omega_j=1$:
$$\mathcal{L}_{\hat\bfpi,\hat\bfeta}( \bfomega, \alpha) =  -\sum_{i = 1}^n \sum_{j = 1}^k \pi_{ij} \log(\omega_j p_{\eta_j}(x_i)) + \lambda H(\bfpi) + \alpha(\sum_{j=1}^k \omega_j - 1)$$ 

Since the function $- \log(\omega)$ is convex, the KKT conditions hold at $(\hat \bfomega, \hat\mu)$, thus $\forall j \in [1,k]$:

\begin{align*}
    \frac{\partial L}{\partial \omega_j} = 0 & \iff     \frac{\partial }{\partial \omega_j} -\sum_{i = 1}^n \sum_{j = 1}^k \hat \pi_{ij} \log(\hat \omega_j p_{\hat \eta_j}(x_i)) + \lambda H(\hat\bfpi) + \hat\alpha(\sum_{j=1}^k \hat\omega_j - 1) = 0 \\
   % &\iff  \frac{\partial }{\partial \omega_j} (\sum_{i=1}^n -\hat \pi_{ij}\log(\hat\omega_j ) + \hat\alpha = 0 \\
    & \iff \sum_i \frac{\hat\pi_{ij}}{\hat\omega_j} = \hat\alpha \\
    & \iff \sum_i \hat\pi_{ij} = \hat\alpha \hat\omega_j,  \; \hat \omega_j \neq 0.
\end{align*}

If $\omega_j = 0 $, since the objective function value is supposed to have a finite value at $(\hat\bfpi, \hat\bfomega,\hat\bfeta)$, then $\pi_{ij} = 0 \enspace \forall i \in [1,n]$.

We know that $\hat\bfomega \in \Sigma_k$, and, since $\upsilon \in \Sigma_n$ and $\hat \bfpi \ones_k = \upsilon$, we also have  $\sum_{ij}\hat \pi_{ij} = 1$. Then, necessarily, $\hat \alpha = 1$ and

\begin{equation*}
    \label{somme_des_omegas}
    \forall j , \; \omega_j = \sum_i \pi_{ij} \enspace .
\end{equation*}

Therefore, $(\hat\bfpi, \hat\bfomega,\hat\bfeta)$ belongs to the set of admissible solutions for problem~\eqref{generalproblem2}. We now denote as $f(\bfpi, \bfomega, \bfeta)$ the objective function minimized in \eqref{generalproblem2} and \eqref{problem}2.
%\newpage
Since $\bfomega$ and $\bfeta$ are defined in the same set for both problems, if $(\hat\bfpi, \hat\bfomega,\hat\bfeta)$ is a coordinatewise minimum, we have:
\begin{align*}
\forall \bfomega \in \Sigma_k, \enspace & f(\hat\bfpi, \hat \bfomega, \hat \bfeta) \leq f(\hat\bfpi, \bfomega, \hat \bfeta) \\
\forall \bfeta \in \Gamma^k, \enspace & f(\hat\bfpi, \hat \bfomega, \hat \bfeta) \leq f(\hat\bfpi, \hat\bfomega,  \bfeta) \\
\forall \bfpi \in \Pi(\upsilon), \enspace & f(\hat\bfpi, \hat \bfomega, \hat \bfeta) \leq f(\bfpi, \hat\bfomega, \hat \bfeta).
\end{align*}
But, since $\Pi(\upsilon,\omega) \subset \Pi(\upsilon)$, we have
$$\forall \bfpi \in \Pi(\upsilon, \omega),\enspace  f(\hat\bfpi, \hat \bfomega, \hat \bfeta) \leq f(\bfpi, \hat\bfomega, \hat \bfeta).$$
Therefore, the solution is also a coordinatewise minimum of problem \eqref{generalproblem2}.

\end{proof}
\section{Convergence analysis}\label{ssec:convergence}

\textit{In this section, we present the detailed proof of Theorem~\ref{thm}}

\begin{definition} Let $f(x_1, \ldots,x_n)$ be a differentiable non necessary convex function. A point $\hat x = (\hat x_1,\ldots,\hat x_n)$ is a stationary point iff $\forall y$,  $\langle \nabla f(\hat x), y - \hat x \rangle \geq 0$. 
\end{definition}

\begin{theorem} Let $(\hat\bfpi, \hat\bfomega,\hat\bfeta)$ be an accumulation point of the algorithm given by the three steps~(9), (10) and (11) of the main paper. If 
\begin{itemize}
    \item (i) $\exists B \subset \Gamma^k$ closed and convex so that all values resulting from step~(11) satisfy  $\bfeta \in B$
    \item (ii) $\forall (i,j), \enspace \log(p_{\eta_j}(x_i))$ is continuously differentiable along $\eta_j$ in the neighborhood of $\hat \bfeta$, and
    \item (iii) $\hat \pi_{ij} > 0$, $\hat \omega_j > 0$
\end{itemize}
then $(\hat\bfpi, \hat\bfomega,\hat\bfeta)$  is a stationary point of the problem~\eqref{problem2} in $\Pi(\bfupsilon) \times \Sigma_k \times B$. Moreover, if $\forall i,j \enspace -\log(p_{\eta_j}(x_i))$ is convex along $\bfeta$, $(\hat\bfpi, \hat\bfomega,\hat\bfeta)$ is a coordinatewise minimum of the function.
\end{theorem}

\begin{proof}
First, the optimization scheme we define is the same as the alternate optimization of problem~\eqref{problem} along $\bfpi$ and $(\bfomega, \bfeta)$. Indeed, for any fixed $\bfpi \in \Pi(\upsilon)$
\begin{align*}
&\inf_{\substack{\bfomega \in \Sigma_k\\ \bfeta \in \Gamma^k}} -\sum_{i = 1}^n \sum_{j = 1}^k \pi_{ij} \log(\omega_j p_{\eta_j}(x_i)) + \lambda H(\bfpi) \\=& \inf_{\bfomega \in \Sigma_k} -\sum_{i = 1}^n \sum_{j = 1}^k \pi_{ij} \log(\omega_j) +  \inf_{\bfeta \in \Gamma^k} -\sum_{i = 1}^n \sum_{j = 1}^k \pi_{ij} \log( p_{\eta_j}(x_i))  + \lambda H(\bfpi) \enspace,
\end{align*}

Then, since all values issued from the optimization step in $\bfeta$ belong to $B$ closed and convex, at each iteration of the algorithm :
\begin{equation*}
\inf_{\bfeta \in \Gamma^k} -\sum_{i = 1}^n \sum_{j = 1}^k \pi_{ij} \log( p_{\eta_j}(x_i)) = \inf_{\bfeta \in B} -\sum_{i = 1}^n \sum_{j = 1}^k \pi_{ij} \log( p_{\eta_j}(x_i))
\end{equation*}
Therefore the optimization scheme corresponds to a block coordinate descent algorithm (BCD) on the 2 variables $\bfpi \in \Pi(\bfupsilon)$ and $(\bfomega, \bfeta) \in \Sigma_k \times B$, where $\Pi(\bfupsilon)$ and $\Sigma_k \times B$ are closed convex sets.\\

Proposition 3 and Corollary 2 in~\cite{grippo2000convergence} state that if $f(x_1,x_2)$ is continuously differentiable in $X_1 \times X_2$ where $X_1$ and $X_2$ are closed convex sets, then any accumulation point $(\bar x_1, \bar x_2)$ of a Block Coordinate Descent (BCD)  algorithm is a stationary point of $f$ in $X_1 \times X_2$. The demonstration only requires $f$ to be continuously differentiable in the neighborhood of $(\bar x_1, \bar x_2)$ and the proposition holds when $X_1$ and $X_2$ are closed convex sets and $f$ is continuously differentiable in the neighborhood of $(\bar x_1, \bar x_2)$.

Therefore,  assumptions (ii) and (iii) guarantee that the function we minimize is differentiable in the neighborhood of $(\hat\bfpi, \hat\bfomega,\hat\bfeta)$ and the results in~\cite{grippo2000convergence}   hold in our case. As a consequence, $(\hat\bfpi, \hat\bfomega,\hat\bfeta)$ is a stationary point of the problem \eqref{problem} in $\Pi(\bfupsilon)\times \Sigma_k \times B$.

Next, if $\bar x = (\bar x_1, \ldots, \bar x_n)$ is a stationary point of $f$ and $\forall i \enspace x_i \mapsto f(x_1,\ldots x_i, \ldots,x_n)$ is convex, then
\begin{align*}
    \forall y_i \in X_i, f(y_i) &\geq f(\bar x) + \langle \nabla_i f(\bar x), y_i - \bar x_i \rangle \\
    &\geq f(\bar x) + \langle \nabla f(\bar x), (0, \ldots, 0, y_i - \bar x_i, 0, \ldots, 0 \rangle \\
    &\geq f(\bar x) \enspace,
\end{align*}
which implies that $\bar x$ is a coordinatewise minimum of $f$. In our context, if $\forall i,j \enspace -\log(p_{\eta_j}(x_i))$ is convex along $\eta_j$, then the optimized function in problem~\eqref{problem} % $-\sum_{i = 1}^n \sum_{j = 1}^k \pi_{ij} \log(\omega_j p_{\eta_j}(x_i)) + \lambda H(\bfpi)$ 
is convex independently in $\bfpi, \bfomega$ and $\bfeta$ and $(\hat\bfpi, \hat\bfomega,\hat\bfeta)$ is a coordinatewise minimum of the function.

\end{proof}

\section{Link with the maximization of the log-likelihood}\label{ssec:likelihood}

%\textit{
In this section, we demonstrate Proposition~\ref{prop2}. We follow a classical proof that shows the increase of the log-likelihood $p(x|\theta)$ along EM iterations, with in our case $\bftheta=(\bfomega, \bfeta)$

Let $q$ be  an arbitrary probability distribution over the unobserved data $\bfpi$, $H$ be the entropy function and recall that the Kullback-Leibler divergence between probability distributions $a$ and $b$ writes $KL(a||b)=\E_a\log(a/b)$. We have

\begin{align*}
\E_q\log(p(\bfx,\bfpi|\bftheta)) - \lambda H(q) &=\E_q\log(p(\bfx,\bfpi|\bftheta))-\lambda \E_q\log(q)\\
&=\E_q\log\left(\frac{p(\bfx,\bfpi|\bftheta)}{p(\bfpi|\bfx,\bftheta)}\right)+\E_q\log\left(\frac{p(\bfpi|\bfx,\bftheta)}{q^{\lambda}}\right)\\
&=\log(p(\bfx|\bftheta))+\lambda\E_q\log\left(\frac{p^{1/\lambda}(\bfpi|\bfx,\bftheta)}{q}\right)\\
&=\log(p(\bfx|\bftheta))-KL\left(q||\frac{p^{1/\lambda}(\bfpi|\bfx,\bftheta)}{||p^{1/\lambda}(\bfpi|\bfx,\bftheta)||_1}\right)+\lambda\log(||p^{1/\lambda}(\bfpi|\bfx,\bftheta)||_1)\enspace .
\end{align*}

The EM algorithm (recovered for $\lambda=1$), consists in maximizing $\E_q\log(p(\bfx,\bfpi|\bftheta)) - \lambda H(q)$ alternatively with respect to $q$ and $\bftheta$. As the KL divergence is non negative and $||p(\bfpi|\bfx,\bftheta)||_1=1$ for $\lambda=1$, this ensures that the log-likelihood $\log(p(\bfx|\bftheta))$ is increased after each step of the EM algorithm. For $\lambda\neq 1$, the situation is different. 

The E-step still corresponds to maximizing the right hand-side with respect to $q$, i.e. $q=\frac{p^{1/\lambda}(\bfpi|\bfx,\bftheta)}{||p^{1/\lambda}(\bfpi|\bfx,\bftheta)||_1}$, which corresponds to the step (9) in the main paper. The M-step corresponds to maximizing $E_q\log(p(\bfx,\bfpi|\bftheta))$ with respect to $\bftheta$, so that it increases the quantity $\log(p(\bfx|\bftheta))+\lambda\log(||p^{1/\lambda}(\bfpi|\bfx,\bftheta)||_1$.  Observing that $\lambda\leq 1$ implies $\log(||p^{1/\lambda}(\bfpi|\bfx,\bftheta)||_1\leq 0$, we conclude that the log-likelihood $\log(p(\bfx|\bftheta))$ is non decreasing if $\lambda\leq 1$.

\section{Hard thresholding with $\lambda\to 0$}\label{app:hard}
\begin{figure}
    \centering
    \includegraphics{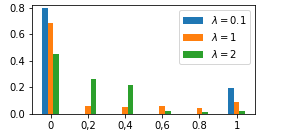}
    \caption{Histograms of  weights $\omega_k$ estimated for different values of $\lambda$. With $\lambda=0.1$, one approximates hard clustering with $99.7\%$ of binary weights.}
    \label{HC}
\end{figure}

We compare in Figure~\ref{HC} the histograms of weights $\bfomega$ obtained for different values $\lambda$. For $\lambda=0.1$,  $99.7\%$ of the estimated weights have binary values $\omega_k\in\{0;1\}$. This illustrates that  hard clustering is recovered for $\lambda\to 0$.

\end{document}